\documentclass{article}
\usepackage{spconf,amsmath,graphicx}

\usepackage{enumitem}
\usepackage{amsmath}
\usepackage{amssymb}
\usepackage{booktabs}
\usepackage{multirow}
\usepackage{xcolor}

\setlist{nosep, leftmargin=14pt}

\usepackage{mwe} 

\usepackage{tikz}
\usepackage{array}

\title{Unsupervised Domain Adaptation with Target-Only Margin Disparity Discrepancy}
\name{Gauthier Miralles$^{\star \dagger}$ \qquad Loïc Le Folgoc$^{\star}$ \qquad Vincent Jugnon$^{\dagger}$ \qquad Pietro Gori$^{\star}$}
 
\address{$^{\star}$ LTCI, Télécom Paris, Institut Polytechnique de Paris, Palaiseau, France \\
    $^{\dagger}$ GE Healthcare, Buc, France}
\begin{document}
\ninept
\maketitle
\begin{abstract}

In interventional radiology, Cone-Beam Computed Tomography (CBCT) is a helpful imaging modality that provides guidance to practicians during minimally invasive procedures. CBCT differs from traditional Computed Tomography (CT) due to its limited reconstructed field of view, specific artefacts, and the intra-arterial administration of contrast medium. While CT benefits from abundant publicly available annotated datasets, interventional CBCT data remain scarce and largely unannotated, with existing datasets focused primarily on radiotherapy applications. To address this limitation, we leverage a proprietary collection of unannotated interventional CBCT scans in conjunction with annotated CT data, employing domain adaptation techniques to bridge the modality gap and enhance liver segmentation performance on CBCT.
We propose a novel unsupervised domain adaptation (UDA) framework based on the formalism of Margin Disparity Discrepancy (MDD), which improves target domain performance through a reformulation of the original MDD optimization framework.
Experimental results on CT and CBCT datasets for liver segmentation demonstrate that our method achieves state-of-the-art performance in UDA, as well as in the few-shot setting.

\end{abstract}
\section{Introduction}
\label{sec:intro}

Interventional radiology enables minimally invasive procedures guided by real-time imaging, reducing patient risk and recovery time. Modern X-ray systems can perform three-dimensional reconstructions using cone-beam computed tomography (CBCT), allowing intraoperative visualization of soft tissues, vessels, and organs. These intraoperative CBCT capabilities make automatic image analysis---such as liver or vessel segmentation---highly desirable to support planning and guidance during procedures.

Despite the clinical value of CBCT, the lack of annotated CBCT datasets severely restricts the development of advanced image-analysis methods. The only publicly available CBCT dataset \cite{cbct_public} originates from radiotherapy rather than interventional procedures. In contrast, large, annotated CT datasets are abundant and publicly accessible \cite{lits}. This disparity motivates the application of \textit{Unsupervised Domain Adaptation (UDA)} to transfer knowledge from labeled CT (source) data to unlabeled CBCT (target) data, thereby reducing annotation costs while preserving accuracy in clinical workflows.

\begin{figure}[t]
\centering
\includegraphics[width=\linewidth]{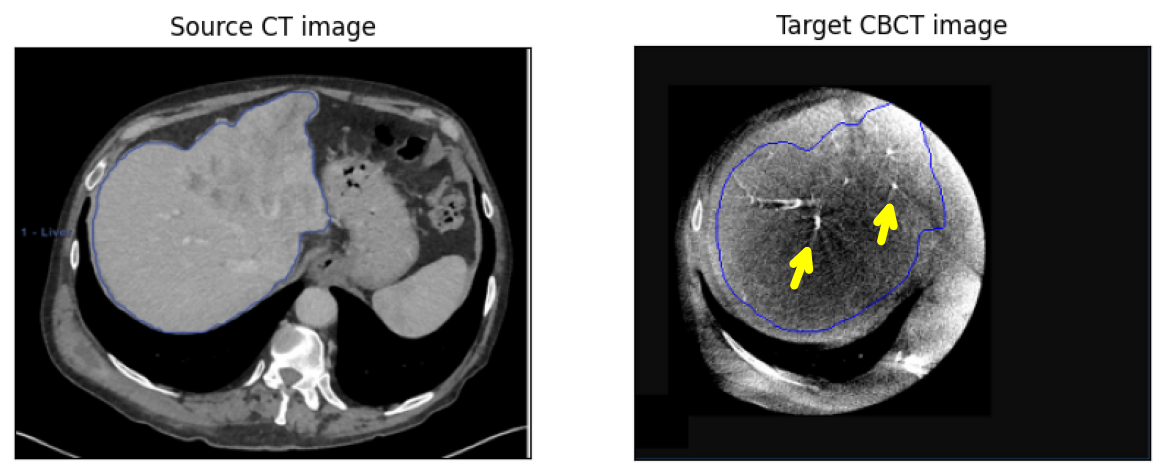}
\caption{Axial slices from abdominal CT (left, source) and CBCT (right, target) of the same patient using identical visualization windows. The liver is shown as a blue overlay. High-intensity regions within the liver caused by intra-arterial
contrast enhancement in CBCT are marked with yellow arrows.}
\label{fig1}
\end{figure}

Domain discrepancies between CT and CBCT arise from several physical and acquisition factors, including scatter, limited dynamic range, and differences in reconstruction geometry (Fig.~\ref{fig1}). These variations lead to both intensity and structural shifts, causing performance degradation when CT-trained models are directly applied to CBCT scans. Addressing these discrepancies is essential for deploying robust, data-driven methods in interventional settings.

\noindent \textbf{Foundation models.} Large-scale foundation models~\cite{DINOv2,SegmentAnything} have recently revived interest in general-purpose approaches for medical imaging. Trained on massive and diverse datasets, these models aim to replace narrow, task-specific pipelines with unified architectures that can transfer broadly across domains and tasks. However, because they are primarily trained on natural images, their generalization to medical modalities is limited. To overcome this, recent studies have proposed fine-tuning foundation models on medical datasets~\cite{SegmentAnything,samed, masam} or applying them in zero-shot scenarios. Nevertheless, they may still underperform on new domains or tasks not encountered during fine-tuning~\cite{general_vs_specific}, highlighting the continued importance of domain adaptation techniques such as UDA.

\noindent \textbf{Unsupervised Domain Adaptation (UDA)} provides a principled framework to mitigate domain gaps between a labeled \emph{source} domain and an unlabeled \emph{target} domain. UDA reduces these discrepancies through feature- or image-level alignment and by transferring task-specific knowledge from the source to the target domain. Existing UDA approaches can be broadly categorized into three main families:

\noindent  \textbf{Feature-alignment.} Feature-alignment approaches reduce the distance between source and target representations through statistical divergences, such as Maximum Mean Discrepancy (MMD) \cite{MMD_1,coral}, or via adversarial learning, as in Domain-Adversarial Neural Networks (DANN) \cite{DANNOrig,DANNSegOrig}. Beyond adversarial learning, contrastive methods have been explored, combining distributional alignment with contrastive objectives \cite{cda, avena}. Adversarial strategies have been extended to segmentation tasks using domain classifiers in U-Net architectures \cite{DAnnUNet}, or by introducing separate or disentangled encoders \cite{adda,dsn}. Among these, the \textit{Margin Disparity Discrepancy (MDD)} framework \cite{MDD_ICML_19} provides a theoretically grounded formulation with demonstrated success in segmentation \cite{MDD_Unet}. Nevertheless, its optimization objective includes a contradictory term in the source domain that may limit effective adaptation, as we empirically show in Sec.~\ref{sec:res}.  

\noindent  \textbf{Image-alignment.} Image-alignment methods adapt the appearance of images between domains using GAN-based style translation. Some use dual generators with shared weights \cite{cogan} or incorporate cycle-consistency to preserve anatomical content \cite{sifa,la_barbera_anatomically_2022}. 
Furthermore, image-alignment methods usually require an identical field of view between source and target images, which is not the case in CT to CBCT UDA.

\noindent  \textbf{Self-training.} Self-training strategies iteratively refine pseudo-labels generated on the target domain, often filtered by confidence thresholds \cite{sgl,pld,cdcl}. More recent formulations combine contrastive learning and prototype regularization to improve pseudo-label consistency \cite{BDCL}. However, these methods can fail under large domain shifts, as their success depends on pseudo-label quality.

\noindent  \textbf{Contributions.} In this work, we propose:  
(1) a novel UDA method inspired by the MDD formalism, modifying its optimization strategy to improve adaptation from CT to CBCT;  
(2) an extension of this approach to the few-shot setting, where a small subset of target annotations is available; and  
(3) a comprehensive evaluation on private abdominal datasets focused on liver segmentation, demonstrating consistent improvements over existing UDA baselines in interventional imaging tasks.

\section{Method}

\begin{figure}[htb]
\centering
\begin{minipage}[b]{1.0\linewidth}
  \centering
  \includegraphics[width=0.95\linewidth]{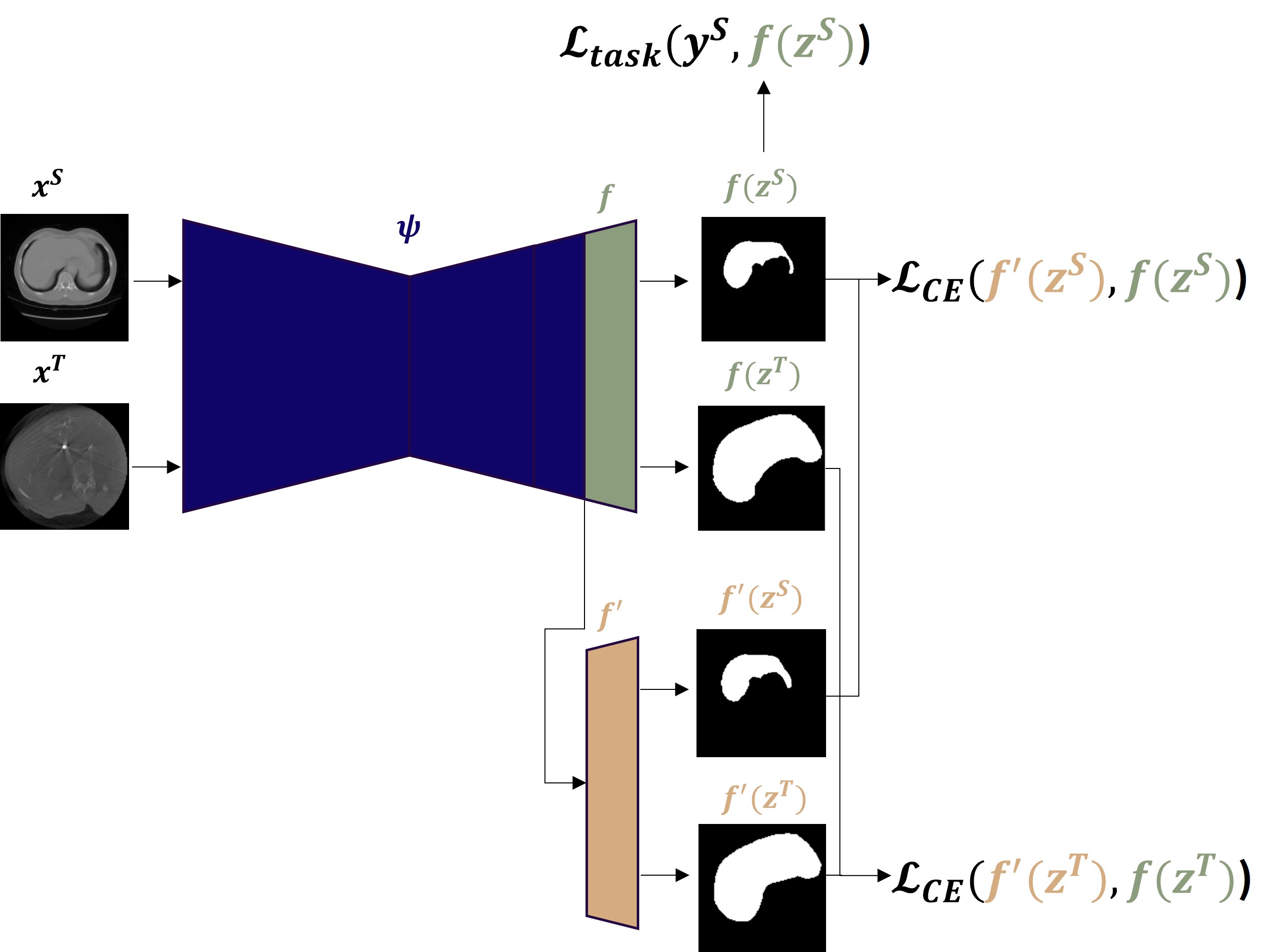}
  \medskip
\end{minipage}
\caption{Our feature-alignment UDA for segmentation. A U-Net is decomposed into a feature extractor $\psi$, and a segmentation head $f$. An adversary $f'$ is built as a duplication of $f$ for adversarial training during UDA. After UDA, the features $z^S = \psi(x^S)$ and $z^T = \psi(x^T)$ are supposed to be domain-invariant, and $f'$ is removed at inference. The loss $\mathcal{L}_{task}$ is the supervision segmentation loss, and $\mathcal{L}_{CE}$ is used for UDA.}
\label{architecture}
\end{figure}

\noindent\textbf{Background: MDD}. We denote $\mathcal{D}^S = (\mathcal{X}^S, \mathcal{Y}, p^S(x, y))$ as the source domain and $\mathcal{D}^T = (\mathcal{X}^T, \mathcal{Y}, p^T(x, y))$ as the target domain. We refer to $\psi$ as the feature extractor, $z \overset{\Delta}{=} \psi (x)$ as the extracted features, $f$ as a classifier/segmentation head and $f'$ as its adversarial classifier/segmentation head. For $(y, \hat{y}) \in [0, 1 ]^{K \times H \times W}$, we denote $\mathcal{L}_{CE}(y, \hat{y})$ as the cross-entropy loss between $y$ and $\hat{y}$ and $\mathcal{L}^{task}$ varies depending on the task: a cross-entropy loss for classification and the sum of the cross-entropy loss and the dice loss for segmentation, for instance. 
MDD \cite{MDD_ICML_19} is a domain adaptation criterion that measures the discrepancy between source $\mathcal{D}^S$ and target $\mathcal{D}^T$ domains by exploiting the margin between decision boundaries of a classifier $f$ and its adversarial counterpart $f'$. The goal is to align source and target feature distributions by minimizing the disparity in classification margins. In \cite{MDD_ICML_19}, authors propose to approximate MDD using cross-entropy losses, to avoid the problem of vanishing gradient. The proposed optimization problem is: 



\begin{align}
\underset{f, \psi}{\text{min}} \;
\underset{f'}{\text{max}} \;
\mathcal{L}^{\text{task}}(f(z^S), y^S)
&+ \alpha\,\mathcal{L}_{CE}(f'(z^T), f(z^T)) \nonumber \\
&- \gamma\,\mathcal{L}_{CE}(f'(z^S), f(z^S))
\label{eq:MDD}
\end{align}

\noindent Howevever, in practice, authors do not optimize the discrepancy losses wrt $f$. The actual optimization problem is thus:


\begin{equation}\label{task_mdd}
\underset{f}{\text{min }} 
\mathcal{L}^{\text{task}}(f(z^S), y^S)
\end{equation}

\begin{align}\label{adversarial_mdd}
\underset{\psi}{\text{min }} \; \underset{f'}{\text{max }} \; 
\mathcal{L}^{\text{task}}(f(z^S), y^S)
&+ \alpha \, \mathcal{L}_{CE}(f'(z^T), f(z^T)) \nonumber \\
&- \tikz[baseline=(X.base)]{
    \node[rectangle, draw=red, thick, rounded corners=2pt, inner sep=1pt](X)
    {$\gamma \, \mathcal{L}_{CE}(f'(z^S), f(z^S))$};
  }
\end{align}

\noindent where $\alpha$ and $\gamma$ are two hyperparameters fixed by the user.
In \cite{MDD_ICML_19} authors claim that, at equilibrium with $\gamma >1$, $\psi$ should extract domain-invariant features.\\
\textbf{Problem: } Looking at Eq. \ref{adversarial_mdd}, we can notice that the last term is contradictory since the feature extractor $\psi$ is optimized so that the discrepancy between $f$ and $f'$ is \textit{maximized} on the source domain, which should not be the case, since the features should reduce the margin between $f$ and $f'$ in both domains.




\noindent\textbf{The proposed Target-Only MDD}. We thus reformulate the adversarial optimization problem of MDD by removing  it and obtaining:
\begin{equation}\label{f_opt}
\underset{f}{\text{min}} \; 
L^{\text{task}}(f(z^S), y^S)
\end{equation}

\begin{equation}\label{f_adv_opt}
\underset{f'}{\text{min}} \; 
\Big[ \mathcal{L}_{CE}(f'(z^S), f(z^S)) - \gamma \, \mathcal{L}_{CE}(f'(z^T), f(z^T)) \Big]
\end{equation}

\begin{align}\label{psi_opt}
\underset{\psi}{\text{min}} \; 
\Big[ & L^{\text{task}}(f(z^S), y^S)
       + \alpha \, \mathcal{L}_{CE}(f'(z^S), f(z^S)) \nonumber \\
       & + \gamma \, \mathcal{L}_{CE}(f'(z^T), f(z^T)) \Big]
\end{align}
where $(\alpha, \gamma) \in \mathbb{R}^+ \times \mathbb{R}^+$ are hyperparameters. As per Eq.~(\ref{f_opt}), $f$ is optimized to solve the  task in the source domain by minimizing $\mathcal{L}_{task}$. Similarly to other MDD-based approaches  \cite{MDD_ICML_19}, \cite{MDD_Unet}, $f'$ is encouraged to predict the same labels as $f$ in source domain and different labels in the target domain, following Eq.~(\ref{f_adv_opt}). However, contrary to MDD, $\psi$ is encouraged to align the predictions of $f$ and $f'$ on both source and target domains. 
\\

\noindent\textbf{Algorithm.} $\psi$ and $f$ are pre-trained in fully-supervised fashion on the source domain by minimizing $\mathcal{L}_{task}$. $f'$ is then initialized with the weights obtained from $f$ as in \cite{MDD_ICML_19}, \cite{MDD_Unet}. We then iteratively optimize Eq.~(\ref{f_opt}), Eq.~(\ref{f_adv_opt}), Eq.~(\ref{psi_opt}) until convergence. \\

\noindent\textbf{Few-shot learning.} While UDA techniques help mitigate domain shift, their performance may still be insufficient for some clinical applications. In practice, clinically acceptable performance may be achievable with the inclusion of a small number of labeled samples from the target domain. The method we propose facilitates this process in a straightforward and effective manner. After aligning feature representations between domains, the source-trained feature extractor $\psi$ and task-specific head $f$ are conserved and the adversary $f'$ is removed. The model, composed of $f \circ \psi$, is then fine-tuned using a small number of expert-annotated samples by minimizing the supervised task loss $\mathcal{L}_{\text{task}}$, enabling reliable performance in the target domain with a limited number of annotated images. 

\section{Experiments and results}\label{sec:res}

\noindent{$\textbf{Dataset.}$}We use a dataset of 573 CBCT and 678 CT 3D volumes consisting of 13,024 CBCT and 15,827 CT 2D slices, respectively.
The data are split at the patient level into training, validation, and test sets, ensuring that no volumes or slices from the same patient appear in more than one split. We use $\frac{2}{3}$ of the patients for training and validation, while the remaining $\frac{1}{3}$ of the patients are reserved for testing.

\noindent{\textbf{Experimental details.}} We conduct two experiments on the task of liver segmentation.
The first experiment is conducted on 2D axial slices extracted from 3D volumes, and all the images are resampled to have a common pixel slice of 1.8mm. The second experiment is conducted directly on the 3D volumes. The same data augmentation strategy was used to increase the training size (i.e., rotation, zoom, random flipping, contrast, noise). We employ a U-Net architecture with five stages 
for all experiment, with 64 channels at stage 1. Each stage consists of a double convolution block, comprising two successive 3×3 convolutional layers followed by LeakyReLU activations. In the encoder, each stage is followed by max pooling for downsampling. In the decoder, each stage is preceded by upsampling and concatenation with the corresponding encoder feature map. An overview of the architecture used in the experiments is displayed in Fig.~\ref{architecture}. The segmentation head $f$ and the adversary $f'$ are composed of the last stage of the U-Net. In both experiments, we use hyperparameters $\alpha = 7.5 \times 10^{-2}$ and $\gamma=3 \times 10^{-1}$. We optimize separately these networks using Adam and different learning rates: $5 \times 10^{-4}$ for $\psi$, $10^{-3}$ for $f$ and $f'$.

\noindent{\textbf{Results.}}
To rigorously evaluate the performance of our method, we compare it with other SOTA methods for UDA \cite{DAnnUNet,MDD_Unet,BDCL, sifa} and two foundation models (i.e., SAM-MED 3D \cite{samed3d}, MA-SAM \cite{masam}). 
For each experiment, we design as Source Only and Target Only a U-Net, with the architecture described above, respectively trained on source domain and target domain. All UDA methods use the same U-Net backbone as previously described.
The performance of SAM-MED 2D/3D is evaluated using 1 and 5 points as prompts, randomly taken in the liver area. We repeat this process 5 times as there might be some variability depending on the localization of these points.

\begin{table}[!ht]
\centering
\caption{2D CT to CBCT segmentation F1 scores for various methods. Best results are highlighted in \textbf{bold}. Standard deviations are reported for experiments repeated 5 times due to randomness in selecting a point for foundation model prompts or randomly choosing slices in few-shot settings.}
\label{tabresults}
\setlength{\tabcolsep}{4pt}
\begin{tabular}{l l c}
\toprule
\textbf{Type} & \textbf{Method} & \textbf{F1 (\%)} $\uparrow$ \\
\midrule

\multirow{6}{*}{\textbf{Fully supervised}}
& Source Only & 54.1 \\
& Target Only (1 vol) & 54.3$\pm$3.7 \\
& Target Only (5 vol) & 68.1$\pm$3.9 \\
& Target Only (20 vol) & 77.4$\pm$0.6 \\
& Target Only (50 vol) & 81.4$\pm$0.5 \\
& Target Only (100\%) & 85.5 \\
\midrule

\multirow{2}{*}{\textbf{Foundation model}}
& SAM-MED 2D \cite{samed} (1 pt) & 44.1$\pm$0.4 \\
& SAM-MED 2D (5 pt) & 67.7$\pm$0.2 \\
\midrule

\textbf{Self-Training} & BDCL \cite{BDCL} & 60.0 \\
\midrule

\multirow{3}{*}{\textbf{Feature Alignment}}
& DANN \cite{DAnnUNet} & 68.3 \\
& MDD \cite{MDD_Unet} & 70.0 \\
& Ours & \textbf{74.4} \\
\midrule

\multirow{4}{*}{\textbf{Few-shot}}
& Ours + 1 vol & 74.5$\pm$1.5 \\
& Ours + 5 vol & 77.2$\pm$1.9 \\
& Ours + 20 vol & 82.4$\pm$0.9 \\
& Ours + 50 vol & 84.6$\pm$0.5 \\
\bottomrule
\end{tabular}
\end{table}

\begin{table}[!ht]
\centering
\caption{3D CT to CBCT segmentation F1 scores for various methods. Best results are highlighted in \textbf{bold}. Standard deviations are reported for experiments repeated 5 times due to randomness in selecting a point for foundation model prompts or randomly choosing scans in few-shot settings.}
\label{tab:results3D_full}
\setlength{\tabcolsep}{4pt}
\begin{tabular}{l l c}
\toprule
\textbf{Type} & \textbf{Method} & \textbf{F1 (\%)} $\uparrow$ \\
\midrule

\multirow{5}{*}{\textbf{Fully supervised}}
& Source Only & 80.1 \\
& Target Only (1 vol) & 61.0$\pm$1.7 \\
& Target Only (5 vol) & 84.7$\pm$0.9 \\
& Target Only (20 vol) & 89.6$\pm$0.6 \\
& Target Only (100\%) & 93.7 \\
\midrule

\multirow{3}{*}{\textbf{Foundation model}}
& SAM-MED 3D \cite{samed3d} (1 pt) & 53.6$\pm$0.3 \\
& SAM-MED 3D (5 pt) & 65.3$\pm$0.1 \\
& MA-SAM \cite{masam} & 61.8 \\
\midrule

\multirow{1}{*}{\textbf{Image Alignment}}
& SIFA \cite{sifa} & 64.7 \\
\midrule

\multirow{1}{*}{\textbf{Self-Training}}
& MAPSeg \cite{MAPSeg} &70.15 \\
\midrule

\multirow{2}{*}{\textbf{Feature Alignment}}
& DANN \cite{DAnnUNet} & 84.6 \\
& Ours & \textbf{86.6} \\
\midrule

\multirow{3}{*}{\textbf{Few-shot}}
& Ours + 1 vol & 87.1$\pm$1.1 \\
& Ours + 5 vol & 90.9$\pm$0.5 \\
& Ours + 20 vol & 91.8$\pm$0.4 \\
\bottomrule
\end{tabular}
\end{table}

\begin{figure*}[htb]
\centering
\includegraphics[width=\textwidth]{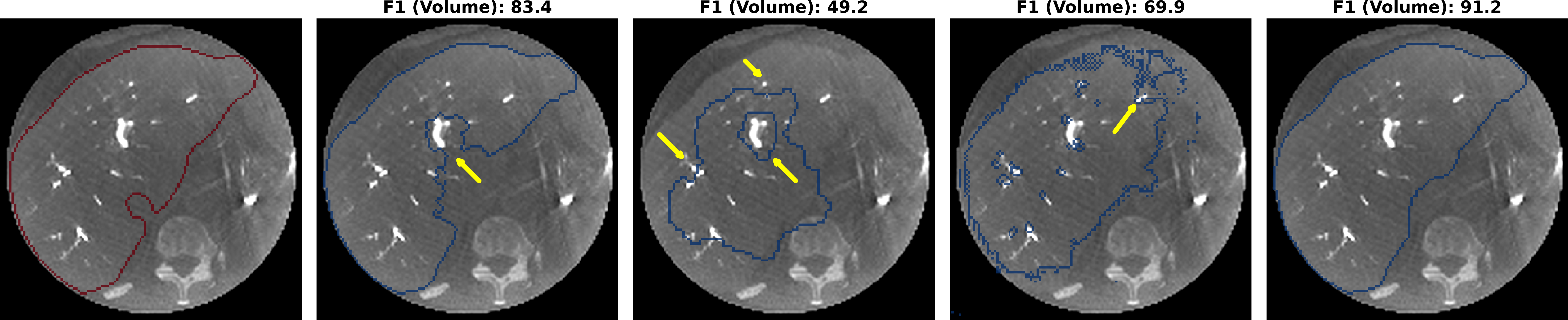}
\vspace{2pt} 
\setlength{\tabcolsep}{0pt} 
\renewcommand{\arraystretch}{1.0} 
\begin{tabular}{*{5}{p{0.20\textwidth}<{\centering}}}
(a) Ground truth & (b) Source only & (c) SAM-MED 3D & (d) MA-SAM & (e) Ours \\
\end{tabular}
\caption{Segmentation results from various 3D methods. Ground truth mask is shown in red, and predicted segmentations in blue. High-intensity regions within the liver that cause the networks to fail are marked with yellow arrows. The same axial slice is displayed for each method with the corresponding prediction overlaid. The SAM-MED 3D prediction was generated using 5 randomly sampled points. The F1 score above each method, reported as a percentage, represents the 3D F1 score computed across the entire volume.}

\label{fig:comparison}
\end{figure*}

\begin{figure}[htb]
\centering
\includegraphics[width=\linewidth]{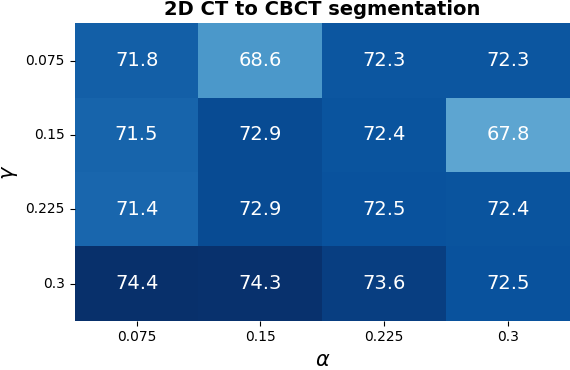}
\caption{Stability Analysis of Hyperparameters in 2D CT to CBCT Unsupervised Domain Adaptation (UDA). This figure reports the F1 scores obtained on the target domain test set for different combinations of the hyperparameters $\alpha$ and $\gamma$, with F1 values presented as percentages. The consistent performance across varying values of $\alpha$ and $\gamma$ confirms the robustness and stability of our method with respect to these hyperparameters.}
\label{tab:stability}
\end{figure}

\noindent \textbf{2D Results.} In Table~\ref{tabresults}, we show that our method outperforms all feature-alignment and self-training SOTA methods, using the Dice coefficient (F1-score) as metric.
Given the significant domain shift in CT to CBCT experiment and considering that initial pseudo-labels are generated using a network pre-trained on the source domain, it is understandable that a self-training approach, such as BDCL \cite{BDCL}, may not be successful. 
Even when provided with 5 prompt points per slice in the liver area, a setting that would be too demanding for clinicians in practice, SAM-MED 2D still performs below our method. Experiments in the few-shot setting demonstrate that our method achieves an F1-score of 84.60\% using only 50 annotated CBCT volumes, which is close to the performance of a model trained from scratch on the full training set of 381 CBCT volumes (85.5\%). 

\noindent \textbf{3D Results.} In Table~\ref{tab:results3D_full}, we present the 3D CT to CBCT segmentation F1 scores obtained with various methods. Transitioning from 2D to 3D considerably enhances overall performance, moving our approach closer to clinically deployable accuracy even without target-domain annotations. In this context, although providing 1 to 5 prompt points in the liver region may be acceptable in a clinical workflow, foundation models such as SAM-MED 3D still fail to achieve satisfactory F1 scores on CBCT segmentation, emphasizing the importance of UDA in this task. Likewise, the recent MA-SAM model, which does not require prompting, performs below our method.
Moreover, SIFA achieves limited performance, likely because image-alignment methods typically assume that source and target images share the same field of view, which does not hold in our dataset. As shown in Fig.~\ref{fig:comparison}, which presents one representative axial slice for each method, the source-only model, SAM-MED 3D and MA-SAM tend to miss high-intensity regions within the liver, likely caused by intra-arterial contrast enhancement. Consequently, the injected vessels lead these networks to under-segment the organ, failing to capture the complete liver boundary, as shown by the yellow arrows. Our approach better captures these high-intensity regions as part of the liver, suggesting that leveraging 3D contextual information with UDA can benefit cross-modality liver segmentation. In the few-shot setting, our UDA model trained without any target annotations already surpasses a target-only model trained with five labeled CBCT volumes. With only five annotations, our method (90.9\%) further outperforms a target-only model trained with twenty labeled scans (89.6\%). Using twenty labeled volumes together with UDA brings the performance close to that of a fully supervised target-only model trained on the full 381-CBCT training dataset.

\textbf{Robustness analysis.} In Fig.~\ref{tab:stability}, we conduct a stability analysis on 2D liver segmentation for CT to CBCT UDA that demonstrates the robustness of our method with respect to the hyperparameters $\alpha$ and $\gamma$. We perform an extensive evaluation across different combinations of these parameters and report the corresponding F1 scores on the target-domain test set. As illustrated in the figure, the performance remains consistently high over a wide range of $\alpha$ and $\gamma$ values, confirming the stability and insensitivity of our method to these hyperparameters. In addition, we evaluate robustness in terms of the variability of the F1 scores measured on the target-domain test set for the 3D CT to CBCT liver segmentation task. A lower spread in these scores reflects a more stable and predictable model. Our method exhibits the lowest standard deviation (9.4\%), indicating substantially less performance fluctuation compared with the source-only model (16.5\%), MA-SAM (18.3\%), and SAM-MED 3D (28.8\%). These results confirm that our approach yields more consistent segmentation performance and is therefore more robust.


\section{Conclusion}
We introduced a novel and generic unsupervised domain adaptation (UDA) method for medical image segmentation, based on a reformulation of the Margin Disparity Discrepancy (MDD) objective. Our approach leverages the formalism of MDD while modifying the optimization process to improve performance on the target domain. To meet the accuracy requirements of clinical applications, we further proposed a few-shot extension of the method. Through extensive experiments on both 2D and 3D CT-to-interventional CBCT datasets, we demonstrated that our approach outperforms state-of-the-art UDA techniques as well as zero-shot transfer using medical foundation models. We show that fine-tuning a model initialized with our UDA requires substantially less annotated target-domain data to achieve performance comparable to training from scratch on a large annotated dataset. \\
\textbf{Limitations.} A limitation of this study is that experiments were conducted only on the liver. While this provides a focused evaluation, the performance of the proposed method on other organs remains to be investigated in future work. \\
\textbf{Future directions.} Future work will focus on extending the proposed method to multiple organs and additional imaging modalities, aiming to evaluate its generalizability and robustness across a wider range of anatomical structures and clinical scenarios.

\section{Compliance with ethical standards}

This study develops an AI algorithm using an anonymized database for which no ethical approval was required.

\section{Acknowledgements}
This work was partially funded by the French Ministry for
Higher Education and Research as part of CIFRE grant No. 2024/0313 and supported by GE Healthcare, where VJ and GM are employed. This work was performed using HPC resources from GENCI-IDRIS (Grant 200319327Z and 2024-A0160615058).

\bibliographystyle{IEEEbib}
\bibliography{strings2,refs2}

\end{document}